\documentclass[runningheads]{llncs}
\usepackage{graphicx}

\usepackage{tikz}
\usepackage{comment}
\usepackage{amsmath,amssymb}
\usepackage{color}
\usepackage{siunitx}

\usepackage[accsupp]{axessibility}

\usepackage[ruled, linesnumbered]{algorithm2e}
\usepackage{floatrow}
\usepackage{caption}
\usepackage{subcaption}
\usepackage{trfsigns}
\usepackage{graphicx}
\usepackage{multirow}
\usepackage{amsmath}
\usepackage{pifont}
\newcommand{\cmark}{\text{\ding{51}}}
\newcommand{\xmark}{\text{\ding{55}}}

\usepackage[pagebackref=true,breaklinks=true,letterpaper=true,colorlinks,bookmarks=false]{hyperref}
\hypersetup{colorlinks=true,linkcolor=red,citecolor=blue,urlcolor=magenta}

\makeatletter
\renewcommand*{\@fnsymbol}[1]{\ensuremath{\ifcase#1\or *\or \dagger\or \ddagger\or
    \mathsection\or \mathparagraph\or \|\or **\or \dagger\dagger
    \or \ddagger\ddagger \else\@ctrerr\fi}}
\makeatother

\begin{document}
\pagestyle{headings}
\mainmatter
\def\ECCVSubNumber{8}  

\title{Multi-modal Depression Estimation based on Sub-attentional Fusion}

\titlerunning{Multi-modal Depression Estimation based on Sub-attentional Fusion}
\author{Ping-Cheng Wei\thanks{The first two authors contribute equally to this work.} \and
Kunyu Peng$^{*}$ \and
Alina Roitberg  \and Kailun Yang \and Jiaming Zhang \and Rainer Stiefelhagen}
\authorrunning{P.-C. Wei \textit{et al.}}
\institute{Institute for Anthropomatics and Robotics, Karlsruhe Institute of Technology\\
\email{pingcheng.wei99@gmail.com, firstname.lastname@kit.edu}\\
\url{https://github.com/PingCheng-Wei/DepressionEstimation}
\thanks{Acknowledgement: The research leading to these results was supported by the SmartAge project sponsored by the Carl Zeiss Stiftung (P2019-01-003; 2021-2026).}
}

\maketitle

\begin{abstract}
Failure to timely diagnose and effectively treat depression leads to over $280$ million people suffering from this psychological disorder worldwide. The information cues of depression can be harvested from diverse heterogeneous resources, \textit{e.g.}, audio, visual, and textual data, raising demand for new effective multi-modal fusion approaches for automatic estimation. In this work, we tackle the task of automatically identifying depression from multi-modal data and introduce a sub-attention mechanism for linking heterogeneous information while leveraging Convolutional Bidirectional LSTM as our backbone. To validate this idea, we conduct extensive experiments on the public DAIC-WOZ benchmark for depression assessment featuring different evaluation modes and taking gender-specific biases into account.
The proposed model yields effective results with $0.89$ precision and $0.70$ F1-score in detecting major depression and $4.92$ MAE in estimating the severity. Our attention-based fusion module consistently outperforms conventional late fusion approaches and achieves competitive performance compared to the previously published depression estimation frameworks, while learning to diagnose the disorder end-to-end and relying on far fewer preprocessing steps.

\end{abstract}

\section{Introduction}
Depression is a common and serious medical condition, negatively impacting the daily lives of ${>}280$ million people according to the World Health Organization (WHO)~\cite{worldhealthorganization2021}.
The severe manifestation of depression is referred to as Major Depressive Disorder (MDD) or Major Depression (MD), which is defined as a mental state of pervasive and persistent low mood, accompanied by the possibility of aversion to activity \cite{bhukya2019major,world2017depression}.

MD is hard to diagnose: the common symptoms, \textit{e.g.}, pessimism, low self-esteem, and cynical behaviour, are more subjective and therefore more difficult to detect compared to most physical illnesses. 
As a consequence, around $33\%$ of patients with depression are not recognized during clinical diagnostic procedures and less than $40\%$ of people with this condition receive proper treatment \cite{jacobi2004prevalence}.
Fortunately, depression is treatable under certain conditions such as early diagnosis \cite{halfin2007depression,kupfer1989advantage}, but making such large-scale diagnostics accessible for the majority of the population will greatly increase the work pressure of psychologists, demonstrating the importance of assistive tools for end-to-end automated estimation of MD.
The release of the Distress Analysis Interview Corpus - Wizard of Oz (DAIC-WOZ) dataset \cite{daicwoz} enabled systematic development and evaluation of learning-based approaches  for depression assessment.
In this dataset, depression cues are learned from audio, visual and textual data, showing promising performance for MD estimation, while audio has been the most common modality in the past work (\textit{e.g.}, DepAudioNet~\cite{ma2016depaudionet}). 
Different cues of MD can be harvested from different types of data: for example, depressed patients tend to speak monotonously, feebly, or anxiously and show less head motion, eye contact, or smiling \cite{fossi1984ethological,waxer1974nonverbal}.
Given the complementary nature of different data sources, multimodality is of key importance for improving automatic depression assessment models, but \textit{how to fuse the information} becomes an important research direction and is the main motivation of our work.

In this work, we introduce a neural network-based multi-modal architecture for MD estimation.
We start by choosing DepAudioNet~\cite{ma2016depaudionet}, one of the best models in the AVEC 2016 challenge~\cite{avec2016}, as our research baseline for MD estimation and propose a novel mechanism for fusing visual, audio, and textual data via attention-based building blocks.
The proposed framework comprises a Convolutaional Bidirectional LSTM (ConvBiLSTM) as our feature extraction backbone and sub-attentional fusion leveraging attention for each individual MD subscore estimation head.
We conduct extensive experiments on the public DAIC-WOZ dataset~\cite{daicwoz}, comparing the proposed sub-attention-based fusion strategy with different score-based late fusion techniques (\textit{e.g.}, addition, concatenation).
We study gender bias in MD estimation and compare our model to previously published approaches, which, in contrast to our work, rely on far stronger feature engineering, preprocessing and data cleaning steps. We further validate our model separately for participant-level and clip-level MD estimation.
Our proposed sub-attentional fusion model outperforms other data fusion techniques and yields competitive performance for MD estimation, while featuring far less feature engineering and data preprocessing than previously published approaches, for the purpose of a more accessible end-to-end automated depression assessment.

\section{Related Work}

\begin{figure}[th]
    \centering
    \includegraphics[width=0.9\linewidth]{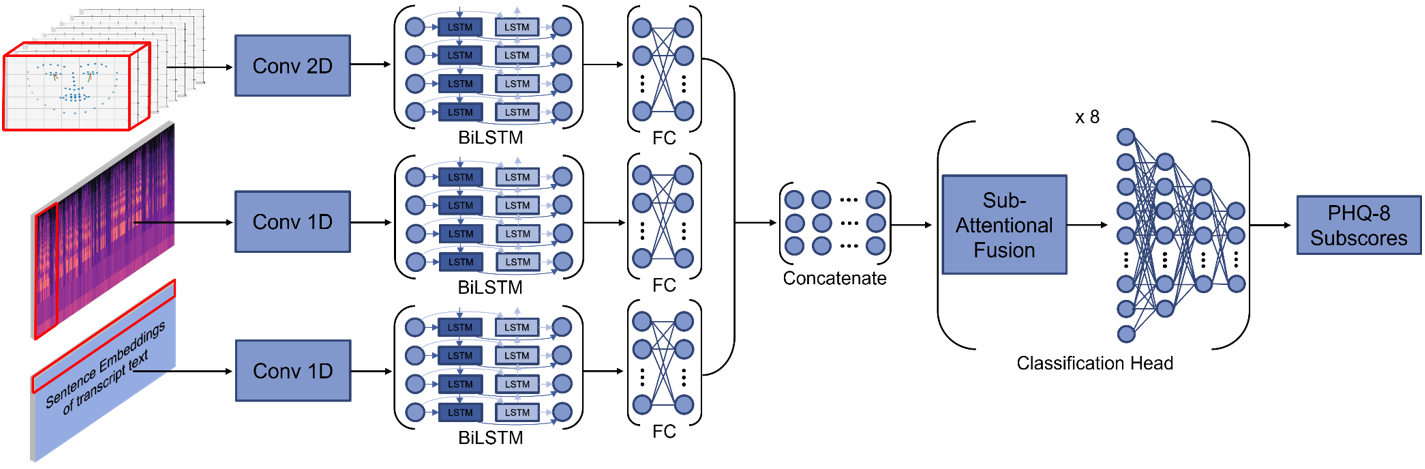}
    \caption[An illustration of the proposed Sub-attentional ConvBiLSTM model structure]
    {An illustration of the proposed Sub-attentional ConvBiLSTM model structure. It has inputs from three different modalities and outputs PHQ-8 Subscores corresponding to the severity of $8$ major depression symptoms. For the backbone, a serial combination of CNN and BiLSTM layers is exploited. After the fully-connected (FC) layers, the extracted features are then concatenated and processed by $8$ different attentional fusion layers, followed by $8$ individual classifiers. 
    This hierarchical structure maximizes the effectiveness of multi-modal features with global- and local attentional fusion. \label{fig:subattenconvbilstm}}
\end{figure}

Benefiting from the rapid development of  data-driven methods, depression estimation through learning-based techniques has attracted remarkable attention. The recently proposed methods targeting  depression estimation can be divided into four groups: text-based, audio-based, vision-based, and multi-modal methods, enabled by the release of the multi-modal depression estimation dataset DAIC-WOZ~\cite{daicwoz}.
According to~\cite{scherer1986vocal}, even slight differences of the psychological state have the potential to cause a noticeable change in the acoustic domain, which makes audio data  a competitive modality to be used  as  input to depression estimation frameworks~\cite{cohn2009detecting,cummins2013diagnosis,dumpala2021estimating,kaya2014eyes,ma2016depaudionet,scherer1986vocal,williamson2013vocal}. 
Mel-cepstral features and formant-frequency tracks are two main arousal representations introduced by Williamson \textit{et al.}~\cite{williamson2013vocal} harvesting discriminative arousal cues from vocal tract resonant frequencies and spectral dynamics.
Scherer \textit{et al.}~\cite{scherer1986vocal} introduced four voice-based features for the psychological distress, which are subsequently classified by Support Vector Machine (SVM).
Since deep learning has gradually took over the pattern recognition field in recent years, remarkable progress has been achieved with the help of deep neural networks instead of hand-crafted feature-based approaches, with an overview of such methods based on audio data provided by He~\textit{et al.}~\cite{he2022deep}.
Excellent performance was reported by Ma~\textit{et al.}~\cite{ma2016depaudionet} who proposed  DepAudioNet, an end-to-end depression estimation model using Convolutional Neural Network (CNN) and Long Short-Term Memory network (LSTM).
Saidi~\textit{et al.}~\cite{saidi2020hybrid} aim at the depression degree estimation fusion via multivariate regression.
Sardari~\textit{et al.}~\cite{sardari2022audio} proposed an audio-based depression detection architecture using a convolutional autoencoder. 
Apart from audio, text data is another popular source of  depression cues. A text-based multi-task Bidirectional Gate Recurrent Unit (BGRU) network for depression estimation is proposed by Dinkel \textit{et al.}~\cite{dinkel2019text}. 
Salimath \textit{et al.}~\cite{salimath2018detecting} proposed a metric to
quantify the depression severity by utilizing negative sentences. Visual cues, which are mainly extracted from facial key points~\cite{chen2021sequential,guo2022automatic,hao2021depression,rathi2019enhanced,zhu2017automated} or raw video data~\cite{akbar2021exploiting,al2018video,pampouchidou2016depression,pampouchidou2017automatic}, also serve for depression estimation by capturing slight facial expression changes. Facial Action Units (FAUs), facial landmarks, head pose and gaze direction are utilized as the CNN input for visual data based approach~\cite{du2019encoding}. Xie \textit{et al.}~\cite{xie2021interpreting} leveraged video data to interpret depression from question-wise long-term video recordings using 3D CNNs.
In order to combine  cues from different modality types, several  fusion architectures are also proposed by the researchers in recent years~\cite{al2018detecting,gong2017topic,joshi2013multimodal,ray2019multi,stepanov2018depression,valstar2016avec}.
A deep multi-modal network for MD assessment is introduced by~\cite{uddin2022deep}. 
He \textit{et al.}~\cite{he2015multimodal} realized a multi-modal depression estimation by combining visual and audio cues. A deeper causal neural network is proposed by Gong \textit{et al.}~\cite{gong2017topic} for the fusion of different modalities.
Drawing inspiration from recent progress in conventional image classification~\cite{dai2021attentional}, we tackle the depression estimation task via cross-modality attention-based fusion.
Furthermore, most of the existing single-modal and multi-modal approaches rely on heavy data preprocessing techniques which violate the end-to-end principle. In contrast to these approaches, we tackled MD assessment by directly using the nearly raw dataset with less data cleaning techniques to train and test the performance of the investigated baselines and our proposed model.

\section{Model}

In this paper, we present a novel deep learning model named Sub-attentional ConvBiLSTM - an attentional multi-modal architecture featuring late fusion of extracted representations from three different modalities.
An overview of our model is provided in Fig.~\ref{fig:subattenconvbilstm}.
Benefiting from its hierarchical structure, Sub-attentional ConvBiLSTM is highly effective in depression estimation while keeping a low computational load, which will be unfolded in Sec.~\ref{sec:experiments}. In addition, two techniques for increasing the performance are introduced, \textit{i.e.}, ``Multi-path Uncertainty-aware Score Distributions Learning (MUSDL) \cite{tang2020uncertainty}'' and ``Sharpness-Aware Minimization (SAM) \cite{foret2020sharpness}''. 
MUSDL is a specific score distribution generation technique for converting each hard-label in the Ground Truth (GT) to a soft-label score distribution for soft decisions. As for SAM, it is a second-order optimization method, which is specifically devised and has been proven~\cite{chen2021vision} to improve the generalization ability of the model, even just training on a small dataset (which is a common case in depression estimation).

\subsection{Sub-attentional ConvBiLSTM}
\label{subsec:subattconvbilstm}
For the input of Sub-attentional ConvBiLSTM, three different feature domains, \textit{i.e.}, log-mel spectrograms (audio), micro-facial expressions (visual), and sentence embeddings (text), have been deployed. 
These inputs will then be processed by each backbone to extract the higher-level representation of each feature. 
The backbone chosen here is based on ``DepAudioNet'' from Ma \textit{et al.}~\cite{ma2016depaudionet}, which was one of the best models in the AVEC 2016~\cite{avec2016} by exploiting a deep-learning-based approach with solely acoustic features for depression detection. 
In general, DepAudioNet uses a serial combination of 1D-CNN layers and LSTM layers with a 2-dimensional input format. We have further improved the model by transforming LSTM layers into Bidirectional LSTM (BiLSTM) layers and enabling the applicability to a 3D input format for visual input by utilizing 2D-CNN layers. 

For the audio and text branch, 1D-CNN layers are first used to provide translation-equivariant responses of a low-level feature map, whose kernel size $k$ is $3$, indicating that several short-term features are captured at these layers. The visual branch, however, uses a 2D-CNN layers with a kernel size $k{=}(72{\times}3)$, as all $72$ key points are perceived as a whole and the kernel slides through the visual data solely along the time-axis, focusing on extracting temporal changes between each frame to provide local attention. Then, batch normalization is performed to regularize the intermediate representation of the features to a standard normal distribution, followed by a nonlinear transformation with the Rectified Linear Unit (ReLU), an activation function defined as $f(x) = max(0, x)$. To further reduce the dimensionality, a max-pooling layer is applied to down-sample the input representation of the feature map. A BiLSTM layer together with an FC layer is stacked at the end of the backbone structure, for an objective of harvesting long-range variability in each modality along the time-axis and retrieving effective features from each branch. After all features from each modality are extracted, they are then concatenated in parallel to form a feature map as input to the subsequent late fusion layer. 

Considering the superiority of weighting fusion methods over the traditional fusion methods \cite{lin2020using} and for more effective deployment of such feature map of multi-modal data, we insert attentional fusion layers inspired by Dai~\textit{et al.}~\cite{dai2021attentional} to realize attentional information interaction between each modality. Details of such a structure as well as its functionality, will be discussed in Sec.~\ref{subsec:attenfusionlayer}. In this Sub-Attentional ConvBiLSTM model, $8$ different attentional fusion layers are exploited, connected with $8$ different output heads, respectively, which correspond to the subclass number of the PHQ-8 Subscores. With this structure, each sub-attentional fusion block will be trained to have the competence in focusing on distinct depression cues from different modalities. For output heads, classifiers are used to predict the PHQ-8 Subscores and the final PHQ-8 Score, the indicator of the severity of depression, as well as the final PHQ-8 Binary, the binary state of having MD, are further derived based on their definitions~\cite{kroenke2009phq}.

With such a network architecture, it is thus expected to not only provide a high-level representation of properties in multimodality, but also comprehensively model the long-term and short-term temporal variabilities of underlying depression cues for precise depression estimation.

\subsection{Attentional Fusion Layer}
\label{subsec:attenfusionlayer}
In the first layer, given a feature map concatenated from extracted features of each modality $\boldsymbol{Y}{\in}\mathbb{R}^{C{\times}H{\times}W}$, it will first be processed by a 2D-CNN layer, which will learn to capture and detect the most critical features to form a new local translation-equivariant response with an identical size of $(C{\times}H{\times}W)$. This response will then be added together with the input feature map $\boldsymbol{Y}$ to form the intermediate feature map $\boldsymbol{X}{\in}\mathbb{R}^{C{\times}H{\times}W}$ as an input for the attentional block in the second layer. $C$ denotes number of channel, which is $1$ in our case, and $H{\times}W$ denotes the size of the feature map. 

\begin{figure}[t]
    \centering
    \includegraphics[width=80mm]{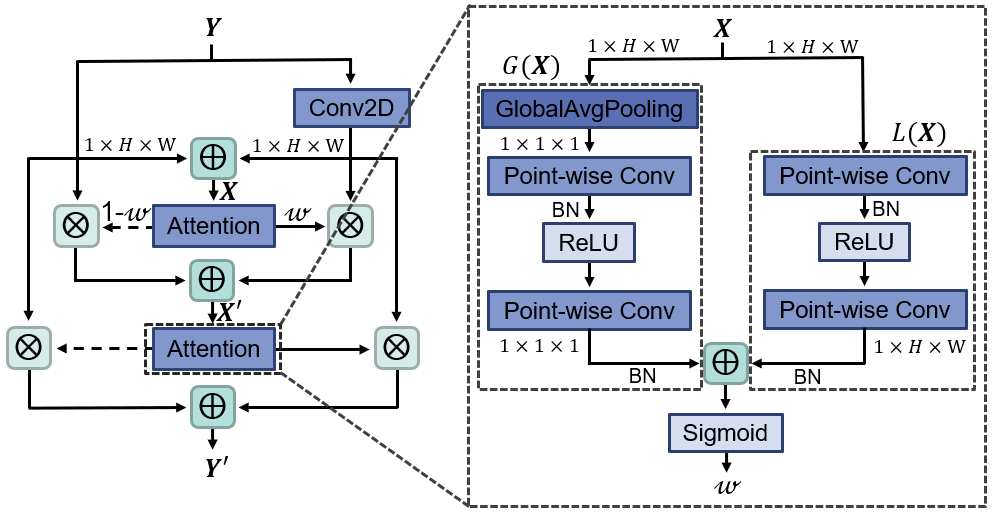}
    \caption[An illustration of the proposed attentional fusion layer structure.]
    {An illustration of the proposed attentional fusion block. A hierarchical structure is built from top to bottom, splitting into three different layers, which is inspired by the work from Dai~\textit{et al.}~\cite{dai2021attentional}. \label{fig:attenfusionlayer}}
\end{figure}

In the second layer, given an intermediate feature map generated from the first layer, the output channel attention weight $w{\in}\mathbb{R}^{C}$  will be computed as:
\begin{equation}
    w = \sigma(G(\boldsymbol{X}) \oplus L(\boldsymbol{X}))),
\end{equation}
which is an aggregation of the global feature attention $G(\boldsymbol{X}){\in}\mathbb{R}^{C}$ and the local channel attention $L(\boldsymbol{X}){\in}\mathbb{R}^{C{\times}H{\times}W}$ transformed through a sigmoid activation function $\sigma$ as shown in the magnified illustration in the Fig.~\ref{fig:attenfusionlayer}. The global feature attention can be obtained via the following equation:
\begin{equation}
    G(\boldsymbol{X}) = BN(W_2 \cdot ReLU(BN(W_1 \cdot g(\boldsymbol{X}))))\,.
\end{equation}
As the name implies, the global feature context of the intermediate feature map will be first extracted through a Global Average Pooling (GAP) block $g(\boldsymbol{X}){=}\frac{1}{H{\times}W} \Sigma_{i=1}^{H} \Sigma_{j=1}^{W} \boldsymbol{X}_{[:,i,j]}$, followed by dimension decreasing and increasing blocks, \textit{i.e.}, $W_1{\in}\mathbb{R}^{\frac{C}{r}\times C}$ and $W_2{\in}\mathbb{R}^{C{\times}\frac{C}{r}}$,  with a Rectified Linear Unit (ReLU) layer in the middle.
$r$ is the channel reduction ratio and both dimension decreasing and increasing blocks are in fact implemented as a point-wise convolution ($PWConv$). After each block, batch normalization ($BN$) is applied. As for the local feature attention, a similar structure can be established via excluding the GAP block $g(\boldsymbol{X})$. Hence, the function could be summarized as:
\begin{equation}
    L(\boldsymbol{X}) = BN(PWConv_2 \cdot ReLU(BN(PWConv_1 \cdot \boldsymbol{X})))\,.
\end{equation}
Moreover, it is noteworthy that the resulting local attentional weight of $L(X)$ has the identical shape $(C{\times}H{\times}W)$ as the input, which can be trained to preserve and highlight the subtle details of depression cues from the intermediate feature map. After the channel attention weight $w$ is derived, the complementary channel attention weight ($1-w$) is also calculated, as denoted with the dashed line in Fig.~\ref{fig:attenfusionlayer}. The refined feature ($RF$) as well as the complementary refined feature ($RF^{c}$) can then be calculated via the following equations:
\begin{equation}
    \begin{split}
        RF &= Conv(\boldsymbol{Y}) \otimes w = Conv(\boldsymbol{Y}) \otimes \sigma(G(\boldsymbol{X}) \oplus L(\boldsymbol{X}))) \,, \\
        RF^{c} &= \boldsymbol{Y} \otimes (1-w) = \boldsymbol{Y} \otimes (1-\sigma(G(\boldsymbol{X}) \oplus L(\boldsymbol{X}))) \,.
    \end{split}
\end{equation}
Finally, the output of the second layer $\boldsymbol{X}'{\in}\mathbb{R}^{C{\times}H{\times}W}$, which is a transitional attentional feature, can be obtained as the summation of both refined features:
\begin{equation}
    \boldsymbol{X}' = RF \oplus RF^{c} = Conv(\boldsymbol{Y}) \otimes w \oplus \boldsymbol{Y} \otimes (1-w)\,.
\end{equation}

In the third layer, the attentional process explained previously will be performed again to further improve and accentuate depressive characteristics in the transitional attentional feature $\boldsymbol{X}'$ of multimodality. Therefore, the ultimate attentional feature fusion output $\boldsymbol{Y}' \in \mathbb{R}^{C \times H \times W}$ can be expressed as:
\begin{equation}
    \boldsymbol{Y}' = Conv(\boldsymbol{Y}) \otimes w' \oplus \boldsymbol{Y} \otimes (1-w') \,,
\end{equation}
where $w'$ is the channel attention weight outputted from $\boldsymbol{X}'$.
Thereby, an adaptive multi-modal interaction is realized, which is beneficial for accurate depression estimation.
In our model, the attentional feature fusion output $\boldsymbol{Y}'$  will then be input to the $8$ classification heads for the PHQ-8 Subscores classification. 

\subsection{MUSDL}
MUSDL stands for Multi-path Uncertainty-aware Score Distributions Learning proposed by Tang~\textit{et al.}~\cite{tang2020uncertainty}. This method converts the hard-label score in GT to a soft-label distribution for soft decisions and has demonstrated effectiveness to solve the intrinsic ambiguity in GT and boost the performance. We flexibly adopt it to reinforce our depression estimation model in harvesting discriminative cues.

Given a classification GT $\boldsymbol{s}_{GT}{\in}\mathbb{N}^{n}$ of an interview clip containing a set of $n$ hard-label scores $s{\in}\mathbb{N}$: $\boldsymbol{s}_{GT}{=}[s_{1},s_{2}, \ldots, s_{n}]$, each score $s$ in the GT will be transformed into a Gaussian-distribution-like soft-label vector $\vec{s}{\in}\mathbb{R}^{m'}$.
It follows $\vec{s}{=}\mathcal{N}(\mu,\,\sigma^{2})$ with a mean $\mu{=}s$ and a standard deviation of $\sigma$, where each hard-label $\{s{\in}\mathbb{N}{\mid}0{\leq}s{<}m\}$ is an integer and the soft-label $\vec{s} = [s'_{1}, s'_{2}, \ldots, s'_{m'}]$ is a discrete set of scores with $\{s'{\in}\mathbb{R}{\mid}0{\leq}s'{\leq}1\}$. Here, $\sigma$ is a hyper-parameter which serves as the level of uncertainty for assessing a clip and $m,\,m'{\in}\mathbb{N}$ denote the class resolution or the number of the classes before and after the soft-label transformation. The transformed ratio $r{\in}\mathbb{R}$ can be derived via $r{=}\frac{m'}{m}$, which should be equal or greater than $1$, indicating an unchanging or expansion of class resolution. The higher the ratio is, the smoother the distribution curve becomes, leading to a better soft-decision strategy performance. In the end, by uniformly discretizing each hard label $s$ in $\boldsymbol{s}_{GT}$ into a normalized soft-label vector $\vec{s}$, a matrix of $n$ Gaussian distributions $\boldsymbol{S}_{GT}^{soft} \in \mathbb{R}^{n \times m'}$ can be obtained. The overall transformation process can be summarized and expressed via:
\newcommand{\vlaplace}[1][]{\mbox
                                {\setlength{\unitlength}{0.1em}%
                                 \begin{picture}(10,20)%
                                 \put(7,2){\circle{4}}%
                                 \put(7,4){\line(0,1){12}}%
                                 \put(7,18){\circle*{4}}%
                                 \put(17,7){#1}
                                 \end{picture}%
                                }%
                            }%
\begin{equation}
	\begin{split}
	\boldsymbol{s}_{GT} &= [s_{1}, s_{2}, \ldots, s_{n}] \\
	&\vlaplace[Label Transformation]  \\
    \boldsymbol{S}_{GT}^{soft} &= [\vec{s}_{1}, \vec{s}_{2}, ..., \vec{s}_{n}] \\
    &= \begin{bmatrix}
    	s'_{1,1} & s'_{1,2} & \cdots & s'_{1,m'} \\
    	s'_{2,1} & s'_{2,2} & \cdots & s'_{2,m'} \\
    	\vdots & \vdots & \ddots & \vdots \\
    	s'_{n,1} & s'_{n,2} & \cdots & s'_{n,m'}
       \end{bmatrix}
       .
    \end{split}
\end{equation}

\begin{figure}[t]
    \centering
    \null
    \hfill
    \begin{subfigure}{.45\textwidth}
        \centering
        \includegraphics[width=\linewidth]{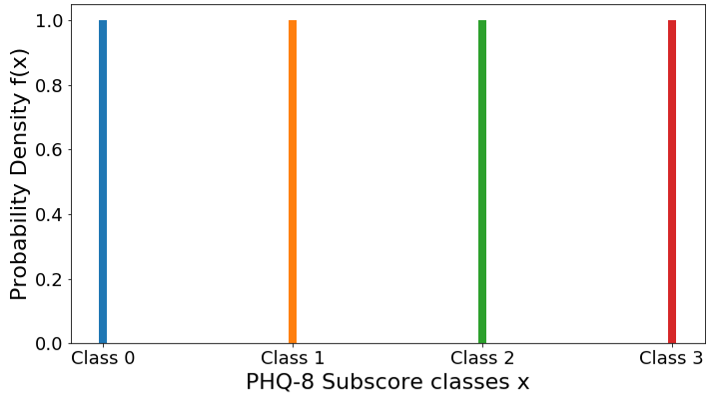}
        \caption[GT hard-label]{GT hard-label.}
        \label{subfig:hardlabel}
    \end{subfigure}
    \hfill
    \begin{subfigure}{.48\textwidth}
        \centering
        \includegraphics[width=\linewidth]{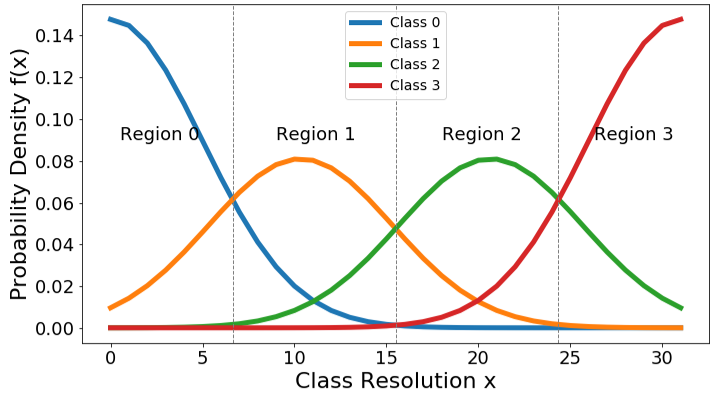}
        \caption[Transformed GT soft-label]{Transformed GT soft-label.}
        \label{subfig:softlabel}
    \end{subfigure}
    \hfill
    \null
    \caption[An overview of label transformation through MUSDL]{An overview of label transformation through MUSDL. (\subref{subfig:hardlabel}) is the $4$ hard-label subclasses in PHQ-8 and (\subref{subfig:softlabel}) is the converted soft-label for each subclass. \label{fig:musdllabel}}
\end{figure}
In this work, n is equal to $8$ and m is $4$ (class 0 to class 3) in accordance with the definition of PHQ-8 Subscores. The standard deviation $\sigma$ is set to $5$ and the transformed ratio $r$ is $8$, indicating that the number of the class is expended from $m = 4$ to $m' = 32$. The final transformed label is illustrated in Fig.~\ref{fig:musdllabel}. One can notice that before the transformation, the hard-label GT of $4$ different classes is given. After the transformation, a probability density function of the normal distribution is generated. Furthermore, during the training stage, all of the $8$ different classification heads are trained to predict the probability between the $4$ different depressive classes of the corresponding subscore with the softmax-function: $\boldsymbol{S}_{pred}^{soft}{=}[\vec{s}_{1,pred}, \vec{s}_{2,pred}, \ldots , \vec{s}_{n,pred}]$. The learning loss is then calculated through pointwise KL divergence between $\boldsymbol{S}_{GT}^{soft}$ and $\boldsymbol{S}_{pred}^{soft}$, which can be computed as:
\begin{equation}
	KL(\boldsymbol{S}_{GT}^{soft} \parallel \boldsymbol{S}_{pred}^{soft}) = \sum^{n}_{i=1} \sum^{m'}_{j=1}\vec{s}_{i,j} \cdot log \frac{\vec{s}_{i,j}}{\vec{s}_{i,j, pred}}\,.
\end{equation}
As for the inference phase, the predicted probability of each class under all PHQ-8 Subscores is derived from the well-trained model and the final assessment $\boldsymbol{s}_{pred} \in \mathbb{N}^{n}$ is obtained by selecting the score with the maximum probability in each subscore, then dividing by the ratio $r$ and rounding down:
\begin{equation}
	\boldsymbol{s}_{pred} = \lfloor \operatorname*{arg\,max}_{\vec{s}_{i,pred}} \{ \vec{s}_{1,pred}, \vec{s}_{2,pred}, \ldots , \vec{s}_{n,pred}\} \slash r \rfloor \, .
\end{equation}

\subsection{SAM}
The problem with the first-order optimization is that even though it minimizes the training loss $L_{train}$, it dismisses the higher-order information such as curvature which correlates with the generalization, leading to a higher generalization error in test loss $L_{test}$ according to \cite{chen2021vision}. Therefore, motivated by Chen~\textit{et al.}~\cite{chen2021vision}, the Sharpness-Aware Minimization (SAM) designed by Foret~\textit{et al.}~\cite{foret2020sharpness}, a second-order optimization technique, is executed to improve the generalization of our model for robust depression estimation in different scenarios. 

Intuitively, SAM seeks to find the weight parameter $w$ of a model whose entire neighbors in the range $\rho$ have low training loss $L_{train}$ compared with other weight parameters, as stated by Chen~\textit{et al.}~\cite{chen2021vision}. This interpretation could be formulated into a minimax decision shown below:
\begin{equation}
	\operatorname*{min}_{w} \space \operatorname*{max}_{\parallel \varepsilon \parallel_{2} \leq \rho} L_{train}(w + \varepsilon) \, ,
\end{equation}
which is a second-order problem. However, due to the complexity of solving the exact inner maximization with the optimum $\varepsilon_{opt}$, Foret~\textit{et al.}~\cite{foret2020sharpness} employ the first-order approximation for better efficiency of calculating the sharpness aware gradient $\hat{\varepsilon}(w)$, which can be structured as:
\begin{equation}
	\begin{split}
	\hat{\varepsilon}(w) &= \operatorname*{arg\,max}_{\parallel \varepsilon \parallel_{2} \leq \rho} L_{train}(w) + \varepsilon^{T} \nabla_{w} L_{train}(w) \\
	&= \rho \nabla_{w} L_{train}(w) \slash \parallel \nabla_{w} L_{train}(w) \parallel_{2} \, .
	\end{split}
\end{equation}
After $\hat{\varepsilon}(w)$ is derived, SAM updates the current weight $w$ based on the $\hat{\varepsilon}(w)$ via the following equation:
\begin{equation}
	w' = \nabla_{w} L_{train}(w)\mid_{w + \hat{\varepsilon}(w)} \, .
\end{equation}

\section{Experiments}
\label{sec:experiments}
In this study, we seek to model sequences of interactions to estimate the depression severity of each individual. Extensive experiments on DAIC-WOZ dataset have been conducted and the overall experimental methodology is to first train each single-modal model, including audio-, visual, and textual data, for the purpose of retrieving weights from effective feature extractors and then applying transfer learning to various multi-modal models. 

\subsection{Dataset}
Distress Analysis Interview Corpus - Wizard of Oz (DAIC-WOZ)
dataset \cite{daicwoz,gratch2014distress} contains clinical interviews of 189 participants designed to support the diagnosis of psychological distress conditions such as anxiety, depression, and post-traumatic stress disorder (PTSD).
During each interview, several data in different format as well as modalities are recorded simultaneously.
However, only the acoustic recordings, facial key points, gaze directions, and transcriptions are chosen in this work, representing 3 different input data domains, namely audio (A), visual (V), and text (T). Moreover, the given GT is an eight-item Patient Health Questionnaire depression scale (PHQ-8), which indicates the severity of depression. A PHQ-8 Score $\geq$ 10 implies that the participant is undergoing a MD \cite{kroenke2009phq}. Although the DAIC-WOZ dataset \cite{daicwoz} abounds in various data types and features, it contains assorted errors and problems, \textit{e.g.}, small-scale dataset, imbalanced dataset, and labeling errors. These issues will potentially sabotage the model performance and mislead the model's attention. Therefore, several techniques are applied to alleviate such burdens, such as sliding window technique, gender balancing (GB), weighted random sampler in PyTorch \cite{NEURIPS2019_9015} etc.

\subsection{Effectiveness of Different Fusion Methods}

During the multi-modal training, we focus on two aspects: the impact pertains to different multimodalities and the effectiveness of the individual fusion approaches.
For multimodality, we conduct experiments based on (1) AVT-modality and (2) AV-modality. As for the fusion approaches, in total, eight different fusion methods have been tested, which could be categorized into the traditional and weighting fusion method as listed below:
\begin{itemize}
    \item Multiplication method: $\boldsymbol{y}^{multi}_{d} = \boldsymbol{x}^{1}_{d} \otimes \boldsymbol{x}^{2}_{d} \otimes \cdots \otimes \boldsymbol{x}^{n}_{d}$,
	\item Concatenation method: $\boldsymbol{y}^{cat}_{n \cdot d} = f^{cat}(\boldsymbol{x}^{1}_{d}, \boldsymbol{x}^{2}_{d}, \cdots, \boldsymbol{x}^{n}_{d})$,
	\item Median method: $\boldsymbol{y}^{median}_{d} = median(\boldsymbol{x}^{1}_{d}, \boldsymbol{x}^{2}_{d}, \cdots, \boldsymbol{x}^{n}_{d})$,
	\item Maximum method: $\boldsymbol{y}^{max}_{d} = max(\boldsymbol{x}^{1}_{d}, \boldsymbol{x}^{2}_{d}, \cdots, \boldsymbol{x}^{n}_{d})$,
	\item Summation method: $\boldsymbol{y}^{sum}_{d} = \boldsymbol{x}^{1}_{d} + \boldsymbol{x}^{2}_{d} + \cdots + \boldsymbol{x}^{n}_{d}$,
	\item Mean method: $\boldsymbol{y}^{mean}_{d} = ( \boldsymbol{x}^{1}_{d} + \boldsymbol{x}^{2}_{d} + \cdots + \boldsymbol{x}^{n}_{d}) \slash n$,
	\item Attentional fusion method,
	\item Sub-attentional fusion method,
\end{itemize}
where $n$ and $d$ are the number and the dimension of extracted feature vectors ($\boldsymbol{x}$).
Furthermore, the attentional fusion method resembles the sub-attentional fusion method. The major difference is the number of attentional fusion layers. 
While the sub-attentional fusion method has individual attentional fusion layer for each of the 8 subclasses, only one single shared attentional layer has been utilized in the attentional fusion method.

The results of different fusion methods are summarized in Table~\ref{tab:multimodality}. On both AV- and AVT-modality, the attention-based reweighting fusion methods generally perform better than the traditional ones with a 1\% accuracy improvement on average.
It indicates that an extra training layer for attentional feature fusion does provide advantages in harvesting deeper underlying depression cues for a better depression estimation.
Moreover, forming the AVT-modality by adding textual data can consistently improve the accuracy of most fusion methods. Our sub-attentional fusion with AVT achieves the best score with 82.65\% of accuracy while showing a satisfied f1-score with 0.65.

\begin{table}[t]
\centering
\caption{Experimental results for multi-modal feature fusion with ConvBiLSTM as backbone. It demonstrates the effectiveness between different fusion methods and modalities.  \label{tab:multimodality}}
\scalebox{0.8}{
\begin{tabular}{ccccc} 
\hline
\multirow{2}{*}{\textbf{~Fusion method~}} & \multicolumn{2}{c}{\textbf{~Accuracy \%~}} & \multicolumn{2}{c}{\textbf{~F1-Score~}}  \\ 
\cline{2-5}
                                          & \textbf{~AV~~}  & \textbf{~~AVT~}            & \textbf{~~AV~} & \textbf{~AVT~}           \\ 
\hline
Multiplication                            & 79.80          & 80.41                     & \textbf{0.63} & 0.58                     \\
~Concatenation~                           & 79.80          & 80.82                     & 0.62          & 0.57                     \\
Median                                    & 80.20          & 82.04                     & \textbf{0.63} & 0.59                     \\
Maximum                                   & 80.82          & 81.22                     & 0.59          & 0.59                     \\
Summation                                 & 81.43          & 81.22                     & 0.61          & 0.63                     \\
Mean                                      & 81.22          & 81.63                     & 0.56          & 0.60                     \\ 
\hline
Attention                                 & \textbf{82.04} & 82.25                     & 0.61          & \textbf{0.66}            \\
Sub-attention                             & \textbf{82.04} & \textbf{82.65}            & 0.58          & 0.65                     \\
\hline
\end{tabular}
}
\end{table}

\subsection{Ablation Studies}
To demonstrate how SAM, BiLSTM, and MUSDL reinforce the performance and have a better understanding of how our models estimate depression between both genders and participants, three following ablation studies have been carried out.

\subsubsection{Effectiveness of applying SAM, BiLSTM, and MUSDL.}
Ablation experiments are conducted regarding the using of SAM, BiLSTM and MUSDL in Table~\ref{tab:sambilstmmusdl}. An incremental performance gain regarding either F1-score or Accuracy is shown by the results demonstrating the efficacy of each individual component of our model, while the combination utilization of all these three techniques shows the best performance regarding audio and visual modality with $76.73\%$ and $79.59\%$ for accuracy, and $0.61$ and $0.61$ for F1-score respectively.

\begin{table}[t]
\centering
\caption{Experimental results of effectiveness before and after applying SAM, BiLSTM, and MUSDL, demonstrated with audio (A) and visual (V) modality. \label{tab:sambilstmmusdl}}
\resizebox{0.55\textwidth}{!}{
\scalebox{0.75}{\begin{tabular}{ccccccc} 
\hline
\multirow{2}{*}{\textbf{~SAM~}} & \multirow{2}{*}{\textbf{~BiLSTM~}} & \multirow{2}{*}{\textbf{~MUSDL~}} & \multicolumn{2}{c}{\textbf{~Accuracy \%~~}} & \multicolumn{2}{c}{\textbf{~~F1-Score ~}}  \\ 
\cline{4-7}
                                &                                    &                                   & \textbf{~A~} & \textbf{~V~}                          & \textbf{~~A~} & \textbf{~V~}                             \\ 
\hline
\xmark                               & \xmark                                  & \xmark                                 & 75.31      & 70.20                                & ~0.53       & 0.52                                             \\
\cmark                               & \xmark                                  & \xmark                                 & 76.12      & 76.73                                & ~0.56       & 0.54                                             \\
\cmark                               & \cmark                                  & \xmark                                 & 76.53      & 76.94                                & ~0.58       & 0.59                                             \\
\cmark                               & \cmark                                  & \cmark                                 & \textbf{76.73}      & \textbf{79.59}                                & \textbf{0.61}       & \textbf{0.61}                                             \\
\hline
\end{tabular}}
}
\end{table}

\subsubsection{Sensitivity of Gender Depression Estimation.}
The purpose of the gender analysis is to dive deep into each modal and comprehend how sensitive each model is in terms of detecting MD between each gender and how significant the Gender Balancing (GB) technique is to suppress the gender bias phenomenon. Therefore, the predicted test results of all clips are categorized into female and male groups, and their results are derived accordingly.
In Table \ref{tab:genderanalysis}, all the results for the gender analysis are summarized with the best score marked in bold.

\begin{table}[t]
\centering
\caption{Experimental results for the analysis of gender bias for depression estimation considering different model structures. \label{tab:genderanalysis}}
\resizebox{1\textwidth}{!}{
\begin{tabular}{cccccccccc} 
\hline
\multirow{2}{*}{\textbf{~Model Name~}} & \multirow{2}{*}{\textbf{~Modality~}} & \multicolumn{4}{c}{\textbf{~Accuracy~\%~}}                                                                                                                                                                                                                & \multicolumn{4}{c}{\textbf{F1-Score}}                                             \\ 
\cline{3-10}
                                       &                                      & \textbf{~Overall~} & \textbf{~Female~} & \textbf{~Male~} & \textbf{~Difference~} & \textbf{~Overall~} & \textbf{~Female~} & \textbf{~Male~} & \textbf{~Difference~}  \\ 
\hline
ConvBiLSTM                             & A (No GB)                            & 70.00                                           & 64.44                                                                                                                            & 77.67                                        & 13.23                 & 0.55               & 0.54              & \textbf{0.59}            & 0.05                   \\
ConvBiLSTM                             & A                                    & 76.73                                           & 79.23                                                                                                                            & 73.30                                        & 5.93                  & 0.61               & \textbf{0.69}              & 0.46            & 0.23                   \\
ConvBiLSTM                             & V                                    & 79.59                                           & 79.93                                                                                                                            & 79.13                                        & 0.80                  & 0.61               & 0.63              & 0.58            & 0.05                   \\
Atten ConvBiLSTM                       & AV                                   & 82.04                                           & 80.63                                                                                                                            & \textbf{83.98}                               & 3.35                  & 0.59               & 0.60              & 0.57            & \textbf{0.03}                   \\
~Sub-atten ConvBiLSTM~                 & AVT                                  & \textbf{82.65}                                  & \textbf{82.39}                                                                                                                   & 82.04                                        & \textbf{0.35}         & \textbf{0.65}               & 0.68              & 0.55            & 0.13                   \\
\hline
\end{tabular}
}
\end{table}

By observing the first two models, which are ConvBiLSTM with and without GB, one can notice a huge reduction of gender accuracy difference of around 7.3\%. This signifies the seriousness of the role that the gender bias phenomenon plays in the acoustic features and how critical it is to handle it during the audio preprocessing stage. Visual features, on the other hand, show no problem of gender bias with a gender accuracy difference of less than 1\%, which is also understandable as one can imagine how challenging it is for a person to distinguish a participant's gender solely based on the 68 3D facial key points.

Furthermore, a decreased tendency of gender accuracy difference in multi-modal model can be discovered, implying that the more different modalities are fused, the lower this acoustic gender bias phenomenon shows up.
This is the fact that by fusing variant data modalities, the model can learn diverse feature from different input sources and balance the gender bias.
Finally, with the Sub-attentional ConvBiLSTM model trained on AVT modality, the lowest gender accuracy difference 0.35\% is achieved, and thus it has the highest sensitivity in gender depression estimation over 82\% accuracy in both genders. The proposed Sub-attentional ConvBiLSTM model also shows the best performance regarding F1-score overall as $0.65$, however, the gender difference regarding F1 score is still a limitation of our model and thereby a future research direction.

\subsubsection{Sensitivity of Participants Depression Estimation.}
\begin{figure}[t]
\centering
\includegraphics[width=0.8\linewidth]{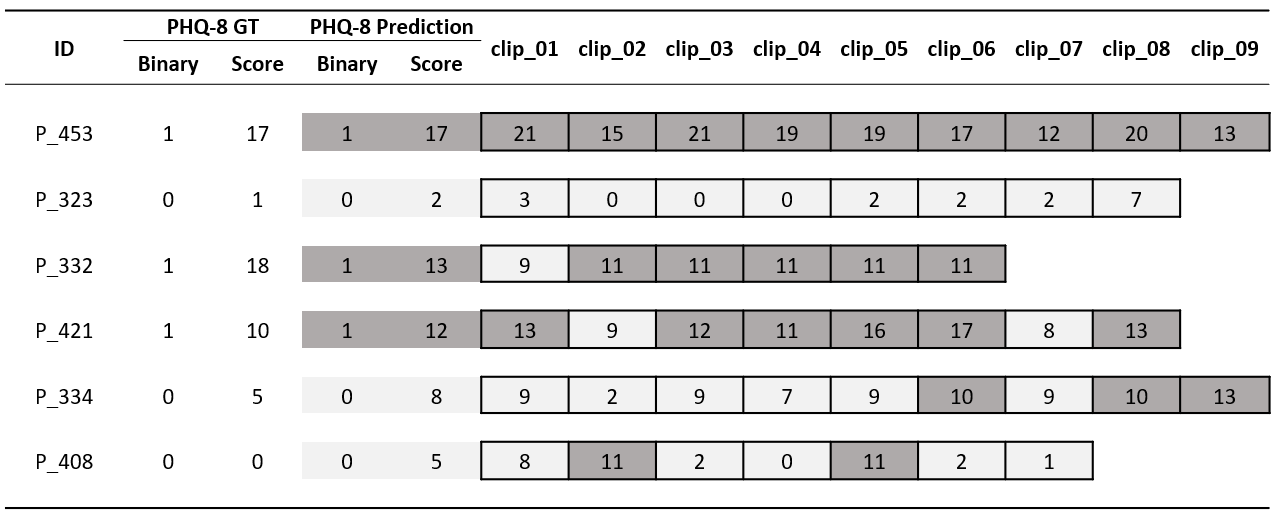}
\caption[A visualization of participant analysis]
{A visualization of the analysis on participant-level. Under each clip, the PHQ-8 Subscore and PHQ-8 Binary are denoted with the value and color inside the block, respectively. Gray background: class 1, White background: class 0. \label{fig:visualrecombine}}
\end{figure}

To further allay the concern regarding the representation of our models since the depression of a participant in GT is diagnosed by the specialist based on a whole interview instead of a clip, the participant analysis is conducted by recombining the clips as well as the predicted scores back into each participant to form the original interview. The final PHQ-8 Score for each participant is then computed as the mean of all clips, and a threshold of 0.5 is set for the final PHQ-8 Binary, meaning that if over 50\% of the clips of the current participant is being classified as depressed by the multi-modal model, it can be concluded that this participant is having MD, and vice versa. An illustration is demonstrated in Fig. \ref{fig:visualrecombine} and the final results are summarized in Table \ref{tab:participantanalysis}.

\begin{table}[t]
\centering
\caption{Experimental results of different model structures for participant-level depression estimation. \label{tab:participantanalysis}}
\resizebox{1\textwidth}{!}{
\begin{tabular}{cccccccc} 
\hline
\multirow{2}{*}{\textbf{~Model Name~}} & \multirow{2}{*}{\textbf{~Modality~}} & \multicolumn{3}{c}{\textbf{~Accuracy~\%~}}                                                                                                                                                                                                               & \multicolumn{3}{c}{\textbf{~F1-Score~}}                                                                                                         \\ 
\cline{3-8}
                                       &                                      & \multicolumn{1}{l}{\textbf{\textbf{~Clipped data~}}} & \multicolumn{1}{l}{\textcolor[rgb]{0.365,0.408,0.475}{\textcolor{black}{\textbf{~Participant-based~}}}\textcolor[rgb]{0.365,0.408,0.475}{}} & \multicolumn{1}{l}{\textbf{\textbf{~Improvement~}}} & \textbf{\textbf{\textbf{\textbf{~Clipped data~}}}} & ~\textbf{\textbf{Participant-based~}} & ~\textbf{\textbf{\textbf{\textbf{Improvement~}}}}  \\ 
\hline
ConvBiLSTM                             & A (No GB)                            & 70.00                                                & 70.21                                                                                                                                       & 0.21                                                & 0.55                                               & 0.53                                  & -0.02                                              \\
ConvBiLSTM                             & A                                    & 76.73                                                & 78.72                                                                                                                                       & 1.99                                                & 0.61                                               & 0.64                                  & 0.03                                               \\
ConvBiLSTM                             & V                                    & 79.59                                                & 78.72                                                                                                                                       & -0.87                                               & 0.61                                               & 0.62                                  & 0.01                                               \\
Atten ConvBiLSTM                       & AV                                   & 82.04                                                & 80.85                                                                                                                                       & -1.19                                               & 0.59                                               & 0.61                                  & 0.02                                               \\
Sub-atten ConvBiLSTM                   & AVT                                  & \textbf{82.65}                                       & \textbf{85.11}                                                                                                                              & \textbf{2.46}                                       & \textbf{0.65}                                      & \textbf{0.70}                         & \textbf{0.05}                                      \\
\hline
\end{tabular}
}
\end{table}

For the normal case, the Sub-attentional model predicts the whole clips from the interview of the participant as class 1 or 0, which is shown in the first two participants in Fig. \ref{fig:visualrecombine}.
For the rest of the specific situations, one can notice a mix of predicted classes for the clips in each interview. 
This is due to the fact of inconsistent expression of depressive symptoms throughout the whole interview despite having MD, which causes non-error mistakes, and some ambiguous clips, which confuse the model. 
This mix, however, can be rectified through the analysis as one can observe from the final binary status of MD in Fig. \ref{fig:visualrecombine}.
Overall, the tolerance between both clipped data and participant-based accuracies is relatively low, less than 3\% according to Table \ref{tab:participantanalysis}. Therefore, it concludes that all of the models, as well as the technique of training on the clipped dataset for depression estimation, are valid and representative. 
Furthermore, one can perceive that there is even a performance improvement of around $2.5\%$ accuracy and $0.05$ F1 score in our best model, Sub-attentional ConvBiLSTM. 

\subsection{Automatic Depression Estimation}

To compare with the state-of-the-art approaches, the following scores are further derived: F1-Score, Precision, Recall, MAE, and RMSE, shown in Table \ref{tab:sotacomparison}. Here, the single- and multi-modal models are both included, along with different analysis approaches, namely clipped data-based as well as participant-based marked with $\dagger$. Moreover, the model reproduction results ($\star$) of the baselines \cite{ma2016depaudionet,al2018detecting}, which are trained on our generated dataset, are also included. The best scores in our methods and previous works are both marked in bold.
One major difference between our approach and prior works is that our approach does not heavily rely on complex feature engineering techniques. We utilize raw input modalities from audio, visual, and text data to realize automatic depression estimation as it is more public-friendly and can potentially improve diagnostic availability, whereas previous works use engineered features such as topic modeling context \cite{gong2017topic}, Question/Answer pair \cite{williamson2016detecting}, vocal tract resonances \cite{williamson2016detecting}, and MFCCs \cite{al2018detecting,valstar2016avec,williamson2016detecting}. However, our models still achieve highly comparable results.

\begin{table}[t]
\centering
\caption{A comparison with the state-of-the-art methods. we assess two outcomes: (1) binary status of having MD and (2) the level of depression severity.}
\resizebox{0.75\textwidth}{!}{
\label{tab:sotacomparison}
\begin{tabular}{l|c|c|ccc|cc} 
\hline
\multicolumn{3}{l|}{\textbf{$\blacksquare$~Comparison of SOTA~}}    & \multicolumn{3}{c|}{\textbf{~PHQ-8 Binary~}}                        & \multicolumn{2}{c}{\textbf{~PHQ-8 Score~}}   \\ 
\cline{1-3}\cline{4-8}
\textbf{Method~}     & \textbf{~Modality~}        &  \textbf{~PL~}                                                                                                   & \textbf{~F1-Score~}    & \textbf{~Precision~}   & \textbf{~Recall~}       & \textbf{~MAE~}         & \textbf{~RMSE~}         \\ 
\hline

\multicolumn{8}{l}{\textbf{Previously Published Works }}                                                                                                                                                                                                                                  \\
Ma \textit{et al.}. \cite{ma2016depaudionet}         & A                          & L                                                                                                 & 0.52                 & 0.35                 & \textbf{1.00}                  & -                    & -                     \\
Valstar \textit{et al.} \cite{valstar2016avec}      & A                          & H                                                                                                & 0.46                 & 0.32                 & 0.86                  & 5.36                 & 6.74                  \\
Williamson \textit{et al.} \cite{williamson2016detecting}    & V                          & H                                                                                                & 0.53                 & -                    & -                     & 5.33                 & 6.45                  \\
Valstar \textit{et al.} \cite{valstar2016avec}      & V                          & H                                                                                                & 0.50                 & 0.60                 & 0.43                  & 5.88                 & 7.13                  \\
Alhanai \textit{et al.} \cite{al2018detecting}       & AT                         & M                                                                                              & \textbf{0.77}                 & \textbf{0.71}                 & 0.83                  & 5.10                 & 6.37                  \\
Valstar \textit{et al.} \cite{valstar2016avec}      & AV                         & H                                                                                                & 0.50                 & 0.60                 & 0.43                  & 5.52                 & 6.62                  \\
Gong \textit{et al.} \cite{gong2017topic}         & AVT                        & H                                                                                                & 0.70                 & -                    & -                     & \textbf{2.77}                 & \textbf{3.54}                  \\
                     &                            &                                                                                                     &                      &                      &                       &                      &                       \\
                     \multicolumn{8}{l}{\textbf{Comparable Baselines} (selected previous works with a data processing pipeline comparable to ours)}                                                                                                                                                                                                                                       \\
$\star$ Ma \textit{et al.} \cite{ma2016depaudionet}          & A                          & ~L~                                                                                                 & 0.48                 & 0.38                 & 0.65                  & -                    & -                     \\
$\star$ Alhanai \textit{et al.} \cite{al2018detecting}       & AT                         & M                                                                                              & 0.44                 & 0.29                 & 0.93                  & 5.92                 & 7.68                  \\
                     & \multicolumn{1}{l|}{}      & \multicolumn{1}{l|}{}                                                                               & \multicolumn{1}{l}{} & \multicolumn{1}{l}{} & \multicolumn{1}{l|}{} & \multicolumn{1}{l}{} & \multicolumn{1}{l}{}  \\
\multicolumn{8}{l}{\textbf{Our Approaches}}                                                                                                                                                                                                                                  \\
ConvBiLSTM           & A                          & L                                                                                                 & 0.61                 & 0.56                 & \textbf{0.66}         & 5.19                 & 6.93                  \\
ConvBiLSTM           & V                          & L                                                                                                 & 0.61                 & 0.64                 & 0.58                  & 6.17                 & 8.06                  \\
Atten ConvBiLSTM     & AV                         & L                                                                                                 & 0.59                 & 0.79                 & 0.47                  & \textbf{4.92}        & \textbf{5.86}         \\
$\dagger$ Atten ConvBiLSTM     & AV                         & L                                                                                                 & 0.61                 & 0.78                 & 0.50                  & 5.06                 & 6.06                  \\
Sub-atten ConvBiLSTM & AVT                        & L                                                                                                 & 0.65                 & 0.73                 & 0.58                  & 4.99                 & 6.67                  \\
$\dagger$ Sub-atten ConvBiLSTM~ & AVT                        & L                                                                                                 & \textbf{0.70}        & \textbf{0.89}        & 0.57                  & 5.04                 & 6.98                  \\
\hline
\end{tabular}

}
\begin{flushleft}
$\dagger$ Participant-based analysis ~ PL: Preprocessing Level ~ H: High ~ M:Medium ~ L: Low \\
$\star$ Model reproduced on the DAIC-WOZ dataset with our preprocessing pipeline ~~
\end{flushleft}
\end{table}

\section{Conclusion}
In this work, we proposed a novel multi-modal deep-learning-based approach, \textit{i.e.}, Sub-attentional ConvBiLSTM, to achieve end-to-end depression estimation while using less prepossessing techniques.
By leveraging multi-modal data with such a hierarchical model structure to capture the short- and long-term temporal as well as spectral features, Sub-attentional ConvBiLSTM has demonstrated great success in harvesting deeper underlying depression cues and thus achieves an exceptional performance with $85.11\%$ for accuracy, $0.89$ for precision, $0.70$ for f1-score, which outperforms our baseline, \textit{i.e.}, DepAudioNet \cite{ma2016depaudionet}, by a large margin.
Furthermore, the proposed gender balancing technique has also been proven to have a strong effect on alleviating gender bias issue in acoustic features.
Finally, the participant-level analysis justifies the efficacy of our model trained on the clipped dataset and leveraging sliding windows during participant-level test. In conclusion, our method has competitive performance with current existed approaches for depression estimation using the knowledge from audio, visual, and text modalities while considering imbalanced, gender bias and small-scale dataset problems, which ensures the efficiency of depression estimation.

\clearpage
\bibliographystyle{splncs04}
\bibliography{egbib}

\begin{thebibliography}{10}
\providecommand{\url}[1]{\texttt{#1}}
\providecommand{\urlprefix}{URL }
\providecommand{\doi}[1]{https://doi.org/#1}

\bibitem{akbar2021exploiting}
Akbar, H., Dewi, S., Rozali, Y.A., Lunanta, L.P., Anwar, N., Anwar, D.:
  Exploiting facial action unit in video for recognizing depression using
  metaheuristic and neural networks. In: ICCSAI (2021)

\bibitem{al2018detecting}
Al~Hanai, T., Ghassemi, M.M., Glass, J.R.: Detecting depression with audio/text
  sequence modeling of interviews. In: Interspeech (2018)

\bibitem{al2018video}
Al~Jazaery, M., Guo, G.: Video-based depression level analysis by encoding deep
  spatiotemporal features. IEEE Transactions on Affective Computing  (2021)

\bibitem{bailey2021gender}
Bailey, A., Plumbley, M.D.: Gender bias in depression detection using audio
  features. In: EUSIPCO (2021)

\bibitem{bhukya2019major}
Bhukya, B.B., Sravanthi, K.: Major depression disorder  (2019)

\bibitem{cer2018universal}
Cer, D., Yang, Y., Kong, S., Hua, N., Limtiaco, N., John, R.S., Constant, N.,
  Guajardo{-}Cespedes, M., Yuan, S., Tar, C., Strope, B., Kurzweil, R.:
  Universal sentence encoder. In: EMNLP (2018)

\bibitem{chen2021sequential}
Chen, Q., Chaturvedi, I., Ji, S., Cambria, E.: Sequential fusion of facial
  appearance and dynamics for depression recognition. Pattern Recognition
  Letters  (2021)

\bibitem{chen2021vision}
Chen, X., Hsieh, C.J., Gong, B.: When vision transformers outperform {ResNets}
  without pre-training or strong data augmentations. In: ICLR (2022)

\bibitem{cohn2009detecting}
Cohn, J.F., Kruez, T.S., Matthews, I., Yang, Y., Nguyen, M.H., Padilla, M.T.,
  Zhou, F., De~la Torre, F.: Detecting depression from facial actions and vocal
  prosody. In: ACII (2009)

\bibitem{cummins2013diagnosis}
Cummins, N., Joshi, J., Dhall, A., Sethu, V., Goecke, R., Epps, J.: Diagnosis
  of depression by behavioural signals: {A} multimodal approach. In: AVEC@ACM
  Multimedia (2013)

\bibitem{dai2021attentional}
Dai, Y., Gieseke, F., Oehmcke, S., Wu, Y., Barnard, K.: Attentional feature
  fusion. In: WACV (2021)

\bibitem{daicwoz}
{DAIC-WOZ Database}: \url{https://dcapswoz.ict.usc.edu/}, accessed Oct. 21,
  2019

\bibitem{dham2017depression}
Dham, S., Sharma, A., Dhall, A.: Depression scale recognition from audio,
  visual and text analysis. arXiv preprint arXiv:1709.05865  (2017)

\bibitem{diagnostic1994statistical}
Diagnostic, A.: Statistical manual of mental disorders (1994)

\bibitem{dinkel2019text}
Dinkel, H., Wu, M., Yu, K.: Text-based depression detection on sparse data.
  arXiv preprint arXiv:1904.05154  (2019)

\bibitem{du2019encoding}
Du, Z., Li, W., Huang, D., Wang, Y.: Encoding visual behaviors with attentive
  temporal convolution for depression prediction. In: FG (2019)

\bibitem{dumpala2021estimating}
Dumpala, S.H., Rempel, S., Dikaios, K., Sajjadian, M., Uher, R., Oore, S.:
  Estimating severity of depression from acoustic features and embeddings of
  natural speech. In: ICASSP (2021)

\bibitem{foret2020sharpness}
Foret, P., Kleiner, A., Mobahi, H., Neyshabur, B.: Sharpness-aware minimization
  for efficiently improving generalization. In: ICLR (2021)

\bibitem{fossi1984ethological}
Fossi, L., Faravelli, C., Paoli, M.: The ethological approach to the assessment
  of depressive disorders. Journal of Nervous and Mental Disease  (1984)

\bibitem{gong2017topic}
Gong, Y., Poellabauer, C.: Topic modeling based multi-modal depression
  detection. In: AVEC@ACM Multimedia (2017)

\bibitem{googleuniversalencoder}
Google: Universal sentence encoder large {V5}. \emph{TensorFlow Hub}. Accessed
  2018 [Online] (2018),
  \url{https://tfhub.dev/google/universal-sentence-encoder-large/5}

\bibitem{gratch2014distress}
Gratch, J., Artstein, R., Lucas, G.M., Stratou, G., Scherer, S., Nazarian, A.,
  Wood, R., Boberg, J., DeVault, D., Marsella, S., Traum, D.R., Rizzo, S.,
  Morency, L.: The distress analysis interview corpus of human and computer
  interviews. In: LREC (2014)

\bibitem{guo2022automatic}
Guo, Y., Zhu, C., Hao, S., Hong, R.: Automatic depression detection via
  learning and fusing features from visual cues. arXiv preprint
  arXiv:2203.00304  (2022)

\bibitem{halfin2007depression}
Halfin, A.: Depression: the benefits of early and appropriate treatment.
  American Journal of Managed Care  (2007)

\bibitem{hao2021depression}
Hao, Y., Cao, Y., Li, B., Rahman, M.: Depression recognition based on text and
  facial expression. In: SPIE (2021)

\bibitem{haque2018measuring}
Haque, A., Guo, M., Miner, A.S., Fei-Fei, L.: Measuring depression symptom
  severity from spoken language and {3D} facial expressions. arXiv preprint
  arXiv:1811.08592  (2018)

\bibitem{he2015multimodal}
He, L., Jiang, D., Sahli, H.: Multimodal depression recognition with dynamic
  visual and audio cues. In: ACII (2015)

\bibitem{he2022deep}
He, L., Niu, M., Tiwari, P., Marttinen, P., Su, R., Jiang, J., Guo, C., Wang,
  H., Ding, S., Wang, Z., Pan, X., Dang, W.: Deep learning for depression
  recognition with audiovisual cues: A review. Information Fusion  (2022)

\bibitem{jacobi2004prevalence}
Jacobi, F., Wittchen, H.U., H{\"o}lting, C., H{\"o}fler, M., Pfister, H.,
  M{\"u}ller, N., Lieb, R.: Prevalence, co-morbidity and correlates of mental
  disorders in the general population: results from the german health interview
  and examination survey {(GHS)}. Psychological Medicine  (2004)

\bibitem{joshi2013multimodal}
Joshi, J., Goecke, R., Alghowinem, S., Dhall, A., Wagner, M., Epps, J., Parker,
  G., Breakspear, M.: Multimodal assistive technologies for depression
  diagnosis and monitoring. Journal on Multimodal User Interfaces  (2013)

\bibitem{kaya2014eyes}
Kaya, H., Salah, A.A.: Eyes whisper depression: A {CCA} based multimodal
  approach. In: ACM Multimedia (2014)

\bibitem{kroenke2002phq}
Kroenke, K., Spitzer, R.L.: The {PHQ-9}: {A} new depression diagnostic and
  severity measure (2002)

\bibitem{kroenke2009phq}
Kroenke, K., Strine, T.W., Spitzer, R.L., Williams, J.B., Berry, J.T., Mokdad,
  A.H.: The {PHQ-8} as a measure of current depression in the general
  population. Journal of Affective Disorders  (2009)

\bibitem{kupfer1989advantage}
Kupfer, D.J., Frank, E., Perel, J.M.: The advantage of early treatment
  intervention in recurrent depression. Archives of General Psychiatry  (1989)

\bibitem{lam2019context}
Lam, G., Dongyan, H., Lin, W.: Context-aware deep learning for multi-modal
  depression detection. In: ICASSP (2019)

\bibitem{lin2020using}
Lin, C.J., Lin, C.H., Jeng, S.Y.: Using feature fusion and parameter
  optimization of dual-input convolutional neural network for face gender
  recognition. Applied Sciences  (2020)

\bibitem{lin2020towards}
Lin, L., Chen, X., Shen, Y., Zhang, L.: Towards automatic depression detection:
  {A} {BiLSTM/1D} {CNN-based} model. Applied Sciences  (2020)

\bibitem{voicechanger}
LingoJam: Male to female voice changer. LingoJam,
  \url{https://lingojam.com/MaletoFemaleVoiceChanger}

\bibitem{ma2016depaudionet}
Ma, X., Yang, H., Chen, Q., Huang, D., Wang, Y.: {DepAudioNet:} {An} efficient
  deep model for audio based depression classification. In: AVEC@ACM Multimedia
  (2016)

\bibitem{world2017depression}
Organization, W.H.: Depression and other common mental disorders: Global health
  estimates. Tech. rep., World Health Organization (2017)

\bibitem{pampouchidou2016depression}
Pampouchidou, A., Simantiraki, O., Fazlollahi, A., Pediaditis, M., Manousos,
  D., Roniotis, A., Giannakakis, G.A., M{\'{e}}riaudeau, F., Simos, P.G.,
  Marias, K., Yang, F., Tsiknakis, M.: Depression assessment by fusing high and
  low level features from audio, video, and text. In: AVEC@ACM Multimedia
  (2016)

\bibitem{pampouchidou2017automatic}
Pampouchidou, A., Simos, P.G., Marias, K., Meriaudeau, F., Yang, F.,
  Pediaditis, M., Tsiknakis, M.: Automatic assessment of depression based on
  visual cues: A systematic review. IEEE Transactions on Affective Computing
  (2019)

\bibitem{NEURIPS2019_9015}
Paszke, A., Gross, S., Massa, F., Lerer, A., Bradbury, J., Chanan, G., Killeen,
  T., Lin, Z., Gimelshein, N., Antiga, L., Desmaison, A., Kopf, A., Yang, E.,
  DeVito, Z., Raison, M., Tejani, A., Chilamkurthy, S., Steiner, B., Fang, L.,
  Bai, J., Chintala, S.: {PyTorch:} {An} imperative style, high-performance
  deep learning library. In: NeurIPS (2019)

\bibitem{rathi2019enhanced}
Rathi, S., Kaur, B., Agrawal, R.: Enhanced depression detection from facial
  cues using univariate feature selection techniques. In: PReMI (2019)

\bibitem{ray2019multi}
Ray, A., Kumar, S., Reddy, R., Mukherjee, P., Garg, R.: Multi-level attention
  network using text, audio and video for depression prediction. In: AVEC@MM
  (2019)

\bibitem{saidi2020hybrid}
Saidi, A., Othman, S.B., Saoud, S.B.: Hybrid {CNN-SVM} classifier for efficient
  depression detection system. In: IC\_ASET (2020)

\bibitem{salimath2018detecting}
Salimath, A.K., Thomas, R.K., Reddy, S.R., Qiao, Y.: Detecting levels of
  depression in text based on metrics. arXiv preprint arXiv:1807.03397  (2018)

\bibitem{sardari2022audio}
Sardari, S., Nakisa, B., Rastgoo, M.N., Eklund, P.: Audio based depression
  detection using convolutional autoencoder. Expert Systems with Applications
  (2022)

\bibitem{scherer1986vocal}
Scherer, K.R.: Vocal affect expression: A review and a model for future
  research  (1986)

\bibitem{song2018human}
Song, S., Shen, L., Valstar, M.: Human behaviour-based automatic depression
  analysis using hand-crafted statistics and deep learned spectral features.
  In: FG (2018)

\bibitem{stepanov2018depression}
Stepanov, E.A., Lathuiliere, S., Chowdhury, S.A., Ghosh, A., Vieriu, R.L.,
  Sebe, N., Riccardi, G.: Depression severity estimation from multiple
  modalities. In: HealthCom (2018)

\bibitem{stevens1937scale}
Stevens, S.S., Volkmann, J., Newman, E.B.: A scale for the measurement of the
  psychological magnitude pitch. The journal of the Acoustical Society of
  America  (1937)

\bibitem{tang2020uncertainty}
Tang, Y., Ni, Z., Zhou, J., Zhang, D., Lu, J., Wu, Y., Zhou, J.:
  Uncertainty-aware score distribution learning for action quality assessment.
  In: CVPR (2020)

\bibitem{uddin2022deep}
Uddin, M.A., Joolee, J.B., Sohn, K.A.: Deep multi-modal network based automated
  depression severity estimation. IEEE Transactions on Affective Computing
  (2022)

\bibitem{avec2016}
Valstar, M., Gratch, J., Schuller, B., Ringeval, F., Lalanne, D.,
  Torres~Torres, M., Scherer, S., Stratou, G., Cowie, R., Pantic, M.: {AVEC
  2016:} {Depression,} mood, and emotion recognition workshop and challenge.
  In: ACM Multimedia (2016)

\bibitem{valstar2016avec}
Valstar, M., Gratch, J., Schuller, B., Ringeval, F., Lalanne, D.,
  Torres~Torres, M., Scherer, S., Stratou, G., Cowie, R., Pantic, M.: {AVEC
  2016:} {Depression,} mood, and emotion recognition workshop and challenge.
  In: AVEC@ACM Multimedia (2016)

\bibitem{velardo2020thesoundofai}
Velardo, V.: Audio signal processing for ml. GitHub (Sep 18, 2020 [Online]),
  \url{https://github.com/musikalkemist/AudioSignalProcessingForML}

\bibitem{waxer1974nonverbal}
Waxer, P.: Nonverbal cues for depression. Journal of Abnormal Psychology
  (1974)

\bibitem{worldhealthorganization2021}
WHO: Depression key facts. World Health Organization (Sep 2021),
  \url{https://www.who.int/news-room/fact-sheets/detail/depression}

\bibitem{williamson2016detecting}
Williamson, J.R., Godoy, E., Cha, M., Schwarzentruber, A., Khorrami, P., Gwon,
  Y., Kung, H.T., Dagli, C., Quatieri, T.F.: Detecting depression using vocal,
  facial and semantic communication cues. In: AVEC@ACM Multimedia (2016)

\bibitem{williamson2013vocal}
Williamson, J.R., Quatieri, T.F., Helfer, B.S., Horwitz, R., Yu, B., Mehta,
  D.D.: Vocal biomarkers of depression based on motor incoordination. In:
  AVEC@ACM Multimedia (2013)

\bibitem{xie2021interpreting}
Xie, W., Liang, L., Lu, Y., Wang, C., Shen, J., Luo, H., Liu, X.: Interpreting
  depression from question-wise long-term video recording of {SDS} evaluation.
  IEEE Journal of Biomedical and Health Informatics  (2022)

\bibitem{zhao2021multi}
Zhao, Y., Liang, Z., Du, J., Zhang, L., Liu, C., Zhao, L.: Multi-head
  attention-based long short-term memory for depression detection from speech.
  Frontiers in Neurorobotics  (2021)

\bibitem{zhu2017automated}
Zhu, Y., Shang, Y., Shao, Z., Guo, G.: Automated depression diagnosis based on
  deep networks to encode facial appearance and dynamics. IEEE Transactions on
  Affective Computing  (2018)

\end{thebibliography}

\appendix
\section{Definition of PHQ-8 System}
\label{sec:defphqsystem}
One of the standardized and validated methods for assessing and diagnosing the severity measure for depressive disorders in large clinical studies is the so-called eight-item Patient Health Questionnaire depression scale (PHQ-8) developed by Kroenke and Spitzer \textit{et al}.~\cite{kroenke2002phq}. The PHQ-8 System consists of 8 of the 9 criteria (also known as PHQ-8 Subscores), on which the DSM-IV diagnosis of depressive disorders is based \cite{diagnostic1994statistical}. These 8 different aspects of depressive criteria are shown in Table \ref{tab:PHQ-8} according to \cite{kroenke2009phq}. 

\begin{table}[ht]
\centering
\caption{An overview of PHQ-8 system, demonstrating the definition and relationship between each others. The 8 subclasses correspond to the $8$ major depression symptoms. \label{tab:PHQ-8}}
\resizebox{\textwidth}{!}{
\begin{tabular}{lcccc} 
\hline
\begin{tabular}[c]{@{}l@{}}\textbf{Over the last 2 weeks, how often have you been }\\\textbf{bothered by any of the following problems?}\end{tabular}                                                                                 & \begin{tabular}[c]{@{}c@{}}\textbf{Not }\\\textbf{ at all }\end{tabular} & \begin{tabular}[c]{@{}c@{}}\textbf{ Several }\\\textbf{ days }\end{tabular} & \begin{tabular}[c]{@{}c@{}}\textbf{ More }\\\textbf{ than }\\\textbf{ half the }\\\textbf{ days }\end{tabular} & \begin{tabular}[c]{@{}c@{}}\textbf{ Nearly }\\\textbf{ every }\\\textbf{ day }\end{tabular}  \\ 
\hline
\multicolumn{5}{l}{\textbf{PHQ-8 Subscores}}                                                                                                                                                                                                                                                                                                                                                                                                                                                                                                                                                     \\
1. Little interest or pleasure in doing things                                                                                                                                                                                        & 0                                                                      & 1                                                                        & 2                                                                                                         & 3                                                                                        \\
2. Feeling down, depressed, or hopeless                                                                                                                                                                                               & 0                                                                      & 1                                                                        & 2                                                                                                         & 3                                                                                        \\
\begin{tabular}[c]{@{}l@{}}3. Trouble falling or staying asleep, \\~ ~ or sleeping too much\end{tabular}                                                                                                                              & 0                                                                      & 1                                                                        & 2                                                                                                         & 3                                                                                        \\
4. Feeling tired or having little energy                                                                                                                                                                                              & 0                                                                      & 1                                                                        & 2                                                                                                         & 3                                                                                        \\
5. Poor appetite or overeating                                                                                                                                                                                                        & 0                                                                      & 1                                                                        & 2                                                                                                         & 3                                                                                        \\
\begin{tabular}[c]{@{}l@{}}6. Feeling bad about yourself \\~ ~ - or that you are a failure \end{tabular}                                                                                      & 0                                                                      & 1                                                                        & 2                                                                                                         & 3                                                                                        \\
\begin{tabular}[c]{@{}l@{}}7. Trouble concentrating on things, such as \\~ ~ ~reading the newspaper or watching television\end{tabular}                                                                                               & 0                                                                      & 1                                                                        & 2                                                                                                         & 3                                                                                        \\
\begin{tabular}[c]{@{}l@{}}8. Moving or speaking so slowly that other people \\~ ~ could have noticed. Or the opposite - being so \\~ ~ fidgety or~restless that you have been moving \\~ ~ around a lot more than usual\end{tabular} & 0                                                                      & 1                                                                        & 2                                                                                                         & 3                                                                                        \\
                                                                                                                                                                                                                                      & \multicolumn{1}{l}{}                                                   & \multicolumn{1}{l}{}                                                     & \multicolumn{1}{l}{}                                                                                      & \multicolumn{1}{l}{}                                                                     \\
\multicolumn{5}{l}{\textbf{\textbf{PHQ-8 Score}}}                                                                                                                                                                                                                                                                                                                                                                                                                                                                                                                                                \\
\multicolumn{5}{l}{~ ~ Total score~ ~ \_\_\_~ =~ \_\_\_~ +~ ......~ +~ \_\_\_~ ~ (sum of all PHQ-8 Subscores, 0 - 24)}                                                                                                                                                                                                                                                                                                                                                                                                                                                                           \\
\multicolumn{5}{l}{}                                                                                                                                                                                                                                                                                                                                                                                                                                                                                                                                                                             \\
\multicolumn{5}{l}{\textbf{\textbf{\textbf{\textbf{PHQ-8 Binary}}}}}                                                                                                                                                                                                                                                                                                                                                                                                                                                                                                                             \\
\multicolumn{5}{l}{~ ~ Final result~ ~ \_\_\_~ =~ 1~ if~ PHQ-8~ Score~ $\geq$ 10~ else~ 0~~}                                                                                                                                                                                                                                                                                                                                                                                                                                                                                                          \\
\hline
\end{tabular}
}
\end{table}

To obtain the PHQ-8 Score, one will be inquired about the number of days in the past 2 weeks one had experienced a particular depressive symptom. Based on the response and the following conversion: 0 to 1 day means "not at all," 2 to 6 days means "several days," 7 to 11 days means "more than half the days," and 12 to 14 days means "nearly every day," the PHQ-8 Subscore for each criterion is acquired by assigning points (0 to 3) to each category, respectively. The results of PHQ-8 Subscores are then summed up to produce a total PHQ-8 Score between 0 to 24 points, from which a binary state of MD is further derived based on a threshold of 10. If PHQ-8 Score $\geq$ 10, it results in an outcome of true classification of having MD, otherwise false. The representation of the depression severity at each numerical range in accord with the PHQ-8 Score is shown in Table \ref{tab:PHQ-8representation}.

\begin{table}[ht]
\centering
\caption{An Illustration of PHQ-8 Score. \label{tab:PHQ-8representation}}
\begin{tabular}{ccc} 
\hline
\textbf{ PHQ-8 Score } & \textbf{ Level of Depressive Symptoms } & \textbf{ State of MD }  \\ 
\hline
0 - 4       & not~significant              & No           \\
5 - 9       & mild                         & No           \\
10 - 14     & moderate                     & Yes          \\
15 - 19     & moderately severe            & Yes          \\
20 - 24     & severe                       & Yes          \\
\hline
\end{tabular}
\end{table}

So far the definition of the PHQ-8 system (GT of the DAIC-WOZ dataset \cite{daicwoz}) has been well explained in-depth. The corresponding underlying relationships among these 3 scores are also established, i.e., PHQ-8 Subscores, PHQ-8 Score, and PHQ-8 Binary, ranging between 0 to 3, 0 to 24, and 0 $\slash$ 1, respectively. Hence, it is conspicuous that 3 different prediction scores can be chosen as the output format of the developed depression estimation architecture and either be considered as a classification predictive modeling problem or a regression predictive modeling problem. A classification head provides an advantage of exact prediction by predicting a discrete class label, which resembles the way PHQ-8 structures, whereas a regression head provides an advantage of minimizing the error in decimal places by predicting a continuous quantity. Therefore, several different variations of prediction for such supervised learning tasks based on the DAIC-WOZ dataset \cite{daicwoz} can be found in the previous automatic depression estimation works. Williamson \textit{et al.} \cite{williamson2016detecting} and Gong \textit{et al.} \cite{gong2017topic} train their model with a regression head by minimizing the RMSE to successfully predict the PHQ-8 Score and further derive the final binary state of MD through the threshold. Ma \textit{et al.}~\cite{ma2016depaudionet} and Bailey \textit{et al.}~\cite{bailey2021gender} regard depression detection as a classification problem and solely predict the binary result of MD of a participant, which is also investigated in other studies \cite{lam2019context,saidi2020hybrid,zhao2021multi}. Alhanai \textit{et al.} \cite{al2018detecting} and Valstar \textit{et al.} \cite{valstar2016avec} design 2 models with 2 different output heads, one with a classification head to model PHQ-8 Binary outcomes and the other with a regression head for multi-class outcomes of PHQ-8 Score. Similar to that, Dham \textit{et al.}~\cite{dham2017depression} also develop 2 models for the classification and regression approach. However, instead of predicting the PHQ-8 Score, the PHQ-8 Subscores were predicted, and the results of the final PHQ-8 Score, as well as the PHQ-8 Binary, are calculated according to the definition. More recently, Haque \textit{et al.}~\cite{haque2018measuring}, Song \textit{et al.}~\cite{song2018human}, and Lin \textit{et al.}~\cite{lin2020towards} deploy a specific criterion function during the training process to fuse the cross-entropy loss and the loss of depression severity assessment since their designed model output with 2 branches, namely a depression classifier for PHQ-8 Binary and a PHQ regression model for PHQ-8 Score.

In this study, in accord with the way PHQ-8 system structures and the consideration of depression estimation as a classification task, \textbf{a classification head, predicting PHQ-8 Subscores, is predominantly exploited in all of the experiments.} PHQ-8 Score and PHQ-8 Binary are then derived through the definition, resembling the method in \cite{dham2017depression}.

\subsection{Potential Problems}
Although the DAIC-WOZ depression database \cite{daicwoz} abounds in various data types and data features, which, to a large extent, benefits numerous research for automatic depression estimation with different data-driven approaches, it has been well reported that the DAIC-WOZ \cite{daicwoz} contains assorted errors and problems, which will not only cause potential difficulties during the model training process but also sabotage the model performance, which, in the worst case, will mislead the model's attention, leading to wholly irrelevant and inapplicable results. Therefore, in this part, these problems as well as our solutions will be discussed.

One of the major challenges in training a shallow or deep depression estimation model with the DAIC-WOZ \cite{daicwoz} lies in the unequal distribution of the dataset, including an uneven sample of depressed and non-depressed participants as well as gender imbalance, which notably appears in acoustic features. It has been widely reported that imbalanced classes in a dataset will greatly affect the performance of the ML model. Moreover, many current benchmarks \cite{bailey2021gender,gong2017topic,ma2016depaudionet,al2018detecting} have shown great adversity of undergoing data imbalance among different levels of depression, which incurs a large bias in the predicted results. Hence, several techniques have been developed to solve this uneven distribution in the dataset. For the inequality of depressed classes with a ratio of 3 to 7, meaning that only 30\% of the participants are being classified as depressed, whereas non-depressed participants constitute about 70\% of the participant, a “weighted random sampler" in PyTorch \cite{NEURIPS2019_9015} is exploited in the data loader to equally load the data from each class of PHQ-8 Score throughout the training despite the PHQ-8 Subscores as predicted scores. This is due to the fact that PHQ-8 Subscores are fixed to each participant and there is not any other way to equalize the number of subclasses while loading the batches based on the clips of the participants. In a compromise, a dynamic weighted loss function is applied, which will dynamically calculate the weight based on the reciprocal of the number derived from the distribution of each subclass per batch and compute the weighted loss accordingly. For the gender imbalance issue, where a total of around 10\% difference in the number of female and male participants is observed, an online software tool \cite{voicechanger} is used to convert each voice of the participants in the recording to the contrary gender for gender balancing, and a new audio dataset is then generated, specifically to train the backbone of the audio branch in our multi-modal model to ameliorate this phenomenon and have the better as well as non-biased capability of extracting depressive characteristic of MD.

Another potential challenge of the DAIC-WOZ \cite{daicwoz} is the scale of the database. There are only 189 participants included in the database, which is a relatively little number of samples compared with the complexity of the depression estimation task. This small-scale dataset not only leads to a hard time to train a representative model but also incurs failures of the generalization ability of a model. This means that the model can encounter at least the following issues: overfitting, underfitting, outliers, sampling bias, missing values, etc. Therefore, to overcome this problem, the "sliding window technique" is applied to segment the interview into $N$ overlapped clips to increase the dataset size, with a window size of $\SI{60}{\s}$ and an overlap size of $\SI{10}{\s}$. With this technique, the scale of the dataset has been expanded tenfold and our model shows a significant performance improvement and stability.

\subsection{Data Preprocessing}
Combining the solutions for the aforementioned problems and techniques for data cleaning as well as data transformation, we propose a framework for preprocessing each data type, \textit{i.e.}, transcriptions, visual data, and acoustic recordings, in the utilization of generating a cleaner and better dataset. This proposed framework, however, is relatively lightweight compared with the data cleaning or feature extraction executed in other existed works~\cite{ma2016depaudionet,williamson2016detecting,al2018detecting} as it solely focuses on normalizing or standardizing the features from the participants and extracting log-mel spectrograms. Previous works, on the other hand, implemented topic modeling or Question/Answer pair, which required an extra building of a preliminary sentence dictionary, manual cleaning of irrelevant sentences, and clustering of the dictionary to groups the sentences with the same topic, and applied sophisticated algorithms to build weighted modeling or extract higher representation of acoustic features such as vocal tract resonances and MFCCs.

\subsubsection{Text Data.} Each transcription record the transcribed conversation of each interview together with the timestamps. Since only the features of the participants are interested in this work, we extract only their sentences along with the corresponding timestamps, which are essential to ensure the alignment of the timeline of other input data domains with text data while conducting late fusion. Furthermore, the pre-trained model of universal sentence encoder (large) from Google \cite{googleuniversalencoder,cer2018universal} is exploited for the generation of sentence embeddings of individually extracted responses which has a Transformer encoder-like architecture and encodes variable-length English sentences to outputs of 512-dimensional arrays.

\subsubsection{Visual Data.} Since both 3D facial key points and gaze directions are provided separately, unnormalized, and need to be reformatted and cropped out the irrelevant parts, we first normalize the the facial key points $\boldsymbol{X}$ to range 0-1 with the following equation: $ \boldsymbol{X}' = a + \frac{(\boldsymbol{X} - \boldsymbol{X}_{min})(b - a)}{(\boldsymbol{X}_{max} - \boldsymbol{X}_{min})}$, where $\boldsymbol{X}, \boldsymbol{X}' \mapsto \mathbb{R}^{3}$ and $a=0, b=1$. The normalized facial key points are then combined with the gaze directions given in 3D unit vectors. Finally, the parts from virtual agent as well as irrelevant interactions are cropped out based on the start-stop time pair extracted from the text preprocessing part.

\subsubsection{Audio Data.} "Log-mel spectrogram" has been chosen in this work for acoustic features as it has been proven to be a more informative and effective audio data format due to the composition of less redundant segments~\cite{bailey2021gender}. We start by filtering and reclipping each given original raw audio signal waveform to have a cleaner and less noisy raw audio signal. The Short-Time Fourier Transformation (STFT) is then applied to the generated clipped raw audio signal for extracting the spectrogram, whose frequency scalar will further be converted to the so-called "Mel scale" designed by Stevens \textit{et al.}~\cite{stevens1937scale}, which resembles the human perception-like frequency scale. Given an N-point discrete-time signal $x[n]$, the STFT-transformed signal $\boldsymbol{X}[k]$ can be calculated as \cite{velardo2020thesoundofai,stevens1937scale}:
\begin{equation}
    \boldsymbol{X}[m,k] = \boldsymbol{STFT}\{x[n]\}= \sum^{L-1}_{n=0} x[n+mH] \cdot w[n] \cdot e^{-j2\pi n\frac{k}{L}}  \qquad  L \leq N \,, 
\end{equation}
where $H = \frac{\text{acoustic sampling rate}}{\text{visual sampling rate}}=\frac{\SI{16}{\kHz}}{\SI{30}{\Hz}}$ denotes the hop size, m denotes the current frame, and $w[n] = \frac{1}{2}[1-\cos{(\frac{2\pi n}{L})}]$ with $0\leq n \leq L$ denotes the L-point Hann window function. The squared magnitude of the $\boldsymbol{X}[k]$ yields the spectrogram representation: $\boldsymbol{Y}[m,k] = \big | \boldsymbol{X}[m,k] \big |^{2}$ and the final log-mel spectrogram (mel filter bank with 80 frequency bins) is computed by processing non-linear transformation to the frequency scalar with the equation below \cite{stevens1937scale}:
\begin{equation}
    f_{mel} = 1127 \cdot ln(1 + \frac{f_{Hz}}{700})\,.
\end{equation}
Lastly, we standardize the extracted log-mel spectrogram $\boldsymbol{Y} \mapsto \mathbb{R}^{2}$ with the following equation: $\boldsymbol{Y}' = \frac{\boldsymbol{Y} - \mu}{\sigma}$, where $\mu$ is the mean of input $\boldsymbol{Y}$ and $\sigma$ is the standard deviation of input $\boldsymbol{Y}$, and acquire the final standardized log-mel spectrogram $\boldsymbol{Y}' \mapsto \mathbb{R}^{2}$.

\end{document}